\documentclass[10pt,twocolumn,letterpaper]{article}
\usepackage{multirow}
\usepackage[accsupp]{axessibility} % Improves PDF readability for those with disabilities.

\usepackage{cvpr}              % To produce the CAMERA-READY version
\usepackage[dvipsnames]{xcolor}

\newcommand{\sh}[1]{{\color{black}#1}}

\definecolor{postechred}{RGB}{0, 0, 0}

\newcommand{\wh}[1]{{\color{black}#1}}

\definecolor{cvprblue}{rgb}{0.21,0.49,0.74}
\usepackage[pagebackref,breaklinks,colorlinks,citecolor=cvprblue]{hyperref}
\usepackage{kotex}  % pdfLaTeX korean supports
\def\paperID{16} % *** Enter the Paper ID here
\def\confName{CVPR}
\def\confYear{2024}

\newcommand{\ds}[1]{\textcolor{black}{#1}}

\title{Burst Image Super-Resolution with Base Frame Selection}

\author{Sanghyun Kim$^{*}$ \hspace{0.8cm} Min Jung Lee$^{*}$ \hspace{0.8cm} Woohyeok Kim  \hspace{0.8cm} Deunsol Jung \hspace{0.8cm} Jaesung Rim  \hspace{0.8cm} \\ Sunghyun Cho \hspace{0.8cm} Minsu Cho \vspace{1.5mm}\\
Pohang University of Science and Technology (POSTECH), South Korea \\
{\tt\small \{sanghyun.kim, minjlee, woohyeok, deunsol.jung, jsrim123, s.cho, mscho\}@postech.ac.kr}
}

\begin{document}
\maketitle
%%%%%%%%% ABSTRACT
\begin{abstract}
Burst image super-resolution has been a topic of active research in recent years due to its ability to obtain a high-resolution image by using complementary information between multiple frames in the burst. 
In this work, we explore using burst shots with non-uniform exposures to confront real-world practical scenarios by introducing a new benchmark dataset, dubbed Non-uniformly Exposed Burst Image (NEBI), that includes the burst frames at varying exposure times to obtain a broader range of irradiance and motion characteristics within a scene.
As burst shots with non-uniform exposures exhibit varying levels of degradation, fusing information of the burst shots into the first frame as a base frame may not result in optimal image quality. 
To address this limitation, we propose a Frame Selection Network~(FSN) for non-uniform scenarios. This network seamlessly integrates into existing super-resolution methods in a plug-and-play manner with low computational costs. 
The comparative analysis reveals the effectiveness of the non-uniform setting for the practical scenario and our FSN on synthetic-/real- NEBI datasets.
\end{abstract}

\let\thefootnote\relax\footnotetext{
$*$ These authors contributed equally to this work.}

%%%%%%%%% BODY TEXT
\section{Introduction}

\begin{figure}[t]
    \centering
    \scalebox{0.95}{
    \includegraphics[width=\columnwidth]{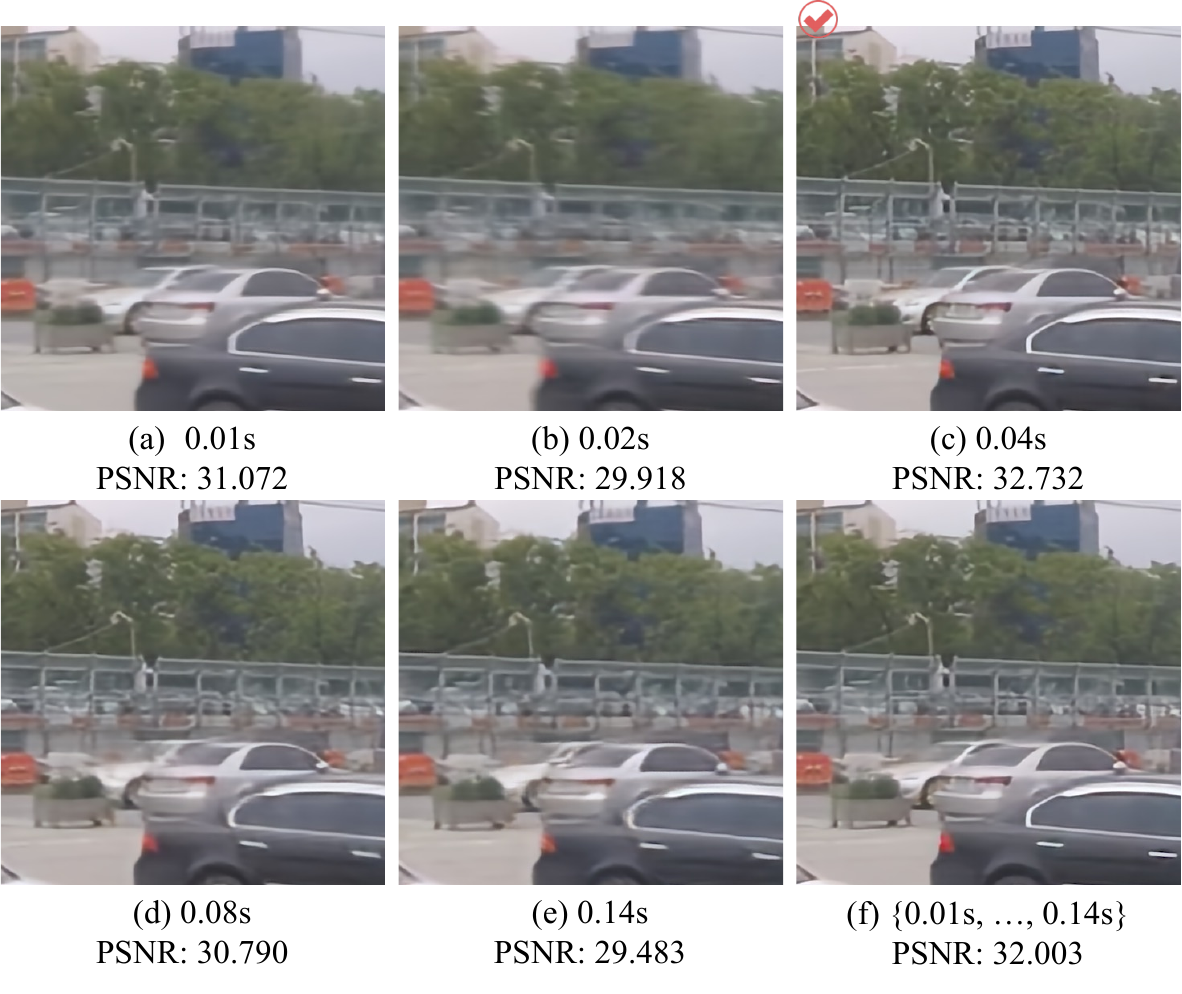}
    } \vspace{-0.4cm}
    \caption{
    Output quality decreases when a BISR model uses burst shots taken at sub-optimal exposure times compared to optimal time~(c).
On the other hand, as shown in (f), the model reconstructs a high-resolution (HR) image as if it were captured using optimally exposed burst shots by harnessing information from non-uniformly exposed burst shots.
}
    \label{fig:uebi_nebi}
    \vspace{-0.6cm}
\end{figure}

Burst image super-resolution~(BISR) is a task to reconstruct a high-resolution~(HR) image with vivid details, utilizing a rapid succession of low-resolution~(LR) frames captured by handheld devices~\cite{ bhat2021deep, hasinoff2016burst,  liba2019handheld, wronski2019handheld}. 
It spans a wide range of computer vision problems such as medical imaging~\cite{isaac2015super,greenspan2009super,mahapatra2019image}, satellite imaging~\cite{shermeyer1812effects,deudon2020highres,cornebise2022open}, object detection~\cite{musunuri2022srodnet,noh2019better,haris2021task} and image generation~\cite{nichol2021glide,rombach2022high,ramesh2022hierarchical}.

In contrast to conventional single-image super-resolution~(SISR) techniques~\cite{dong2015image, zamir2020learning, zamir2021multi, zhang2020residual} where the inherent challenges of ill-posed problems constrain their performance, BISR demonstrates the ability to achieve high-quality images in terms of signal-to-noise ratio (SNR) through the utilization of multiple images for restoration and enhancement.
Despite the advantages conferred by utilizing multiple input images for super-resolution, the successive frames are usually acquired with uniform exposure times~\cite{hasinoff2016burst, bhat2021deep}.
This may induce inferior-quality outputs of the enhancement network utilizing the burst shots captured at the suboptimal exposure time due to inadequate camera noise and motion blur present in those shots.
As shown in Fig.~\ref{fig:uebi_nebi}, the output quality significantly declines when utilizing burst shots acquired far from the optimal exposure time in the uniformly exposed setting.
This approach becomes impractical in real-world scenarios where ascertaining the optimal exposure time is challenging.
Meanwhile, we observe that the model utilizing burst shots acquired with non-uniform exposures can reconstruct an HR image as if it uses the optimal exposure time.

Motivated by this observation, we explore using burst shots with non-uniform exposures for super-resolution in real-world scenarios, introducing a novel benchmark named Non-uniformly Exposed Burst Image (NEBI).
Unlike existing burst image super-resolution benchmarks~\cite{hasinoff2016burst, bhat2021deep},
our approach emphasizes non-uniform burst imaging, capturing images at varying exposure times.
This enables the acquisition of images with a broader range of irradiance and motion characteristics within a scene for practical scenarios.

\begin{figure}[t]
    \centering
    \scalebox{0.90}{
    \includegraphics[width=\columnwidth]{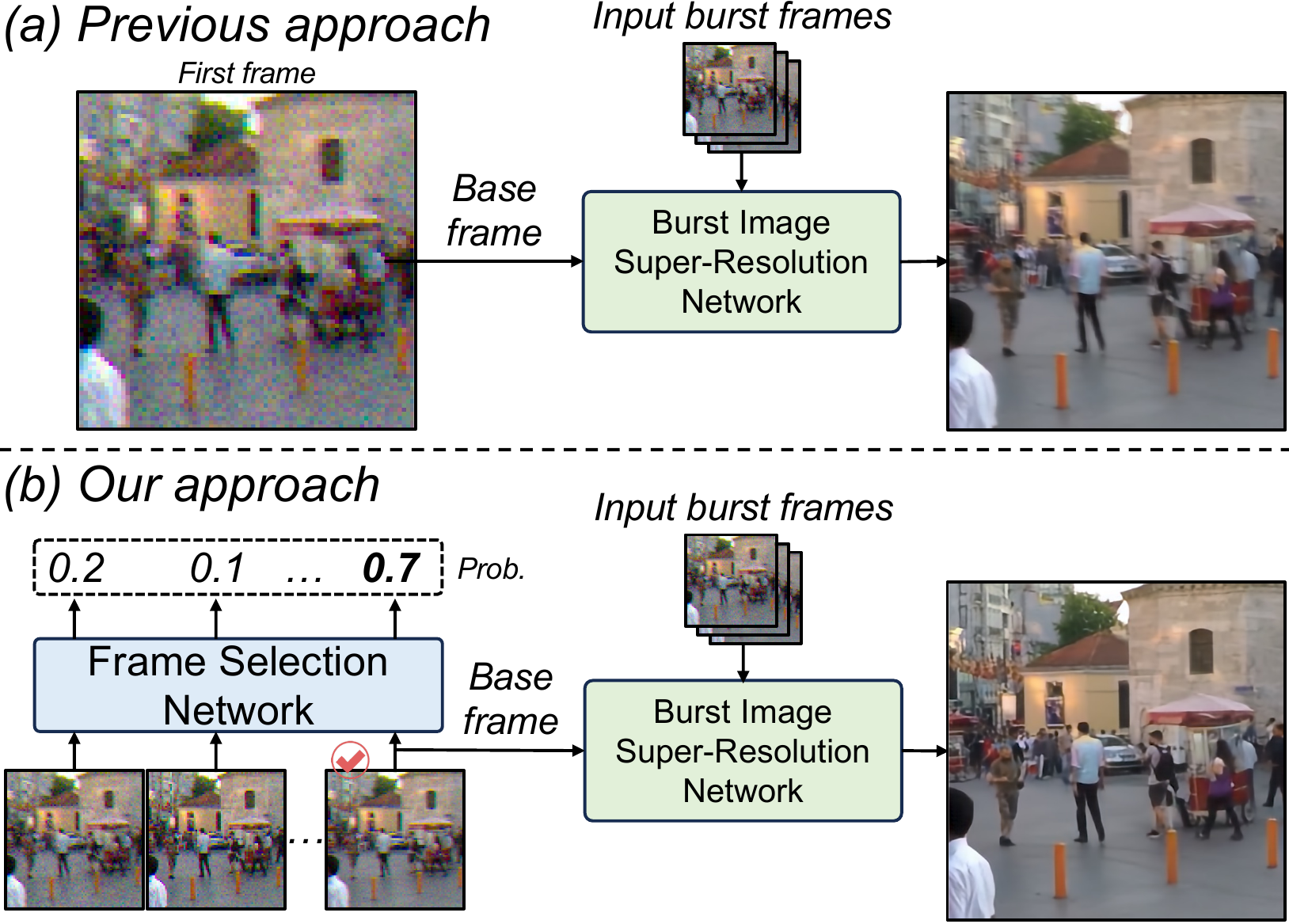}
    } \vspace{-0.2cm}
    \caption{ 
    (a) Previous methods~\cite{bhat2021deep,bhat2021deep_repara,dudhane2022burst} fix the base frame to the first frame, which often suffers from degradation~(\textit{e.g.}, noise, blur), resulting in poor image quality.
    (b) Our approach dynamically selects the base frame using a frame selection network, enhancing the quality of the final output image. This adaptive strategy helps the model generate a cleaner final output image.
    }
    \label{fig:comp_archi}
    \vspace{-0.6cm}
\end{figure}

In contrast to burst shots with a uniform exposure, where all frames exhibit similar levels of degradation (\textit{e.g.,} blur, noise), burst shots with non-uniform exposures present varying degrees of degradation.
The degradation of the base frame, which merges information from subsequent frames, may interfere with the feature alignment and fusion processes between the base frame and the other frames, resulting in a noticeable decline in the output image quality.
However, existing methods \cite{bhat2021deep, dudhane2022burst, bhat2021deep_repara} assume that the first frame of the input bursts serves as the base frame without considering the importance of the base frame on the image restoration.
As shown in Fig.~\ref{fig:comp_archi}~(a), the quality of the output is decreased when the first frame is always used as the base frame. 
To address this limitation, we also introduce a Frame Selection Network~(FSN) that automatically discerns the most suitable base frame as shown in Fig.~\ref{fig:comp_archi}~(b).
Our FSN effectively eliminates the presumption that the first frame is invariably the optimal choice, improving the overall image quality.
Furthermore, our FSN seamlessly integrates into prevailing burst enhancement networks, offering plug-and-play compatibility.
Our proposed method achieves a noteworthy enhancement in quality compared to existing stand-alone super-resolution methods, as demonstrated by superior performance in perceptual metrics on synthetic and real NEBI datasets.

The contributions of our paper are threefold:
\begin{itemize}
\item[$\bullet$] We employ non-uniformly exposed burst shots for practical scenarios where optimal exposure times are unknown in the super-resolution task.
\item[$\bullet$] 
We introduce the Non-uniformly Exposed Burst Image (NEBI) benchmark, featuring varied exposure times.
\item[$\bullet$] We propose a novel Frame Selection Network (FSN) that selects the most suitable base frame for BISR in a plug-and-play manner.

\end{itemize}

\section{Related Work}

% \noindent \textbf{Burst Image Enhancement and Restoration.}
\noindent \textbf{Burst image enhancement.}
Burst denoising~\cite{dahary2021digital, godard2018deep, liu2014fast} and burst deblurring~\cite{aittala2018burst,wieschollek2017end} have recently been studied with the development of handheld cameras.
In addition to burst denoising, burst images have been utilized for low-light denoising to enhance images captured in very dark conditions~\cite{chen2018learning, hasinoff2016burst, karadeniz2021burst, liba2019handheld}.
For the burst image super-resolution, a seminal classical paper~\cite{tsai1984multiframe} proposes a restoration model using multiple frames with known translation between frames at the frequency domain.
Many following approaches~\cite{elad1997restoration,farsiu2004fast, hardie2007fast,kohler2018multi,takeda2007kernel} have extended it to the spatial domain to solve burst image super-resolution.
Current learning-based methods~\cite{deudon2020highres, lecouat2022high, lecouat2021lucas} formulate the burst-frame super-resolution joint learning to align burst frames.
DBSR~\cite{bhat2021deep} proposes a burst image enhancement benchmark and sets the baseline model to predict a DSLR-quality image from input bursts captured by a handheld device, and Deep-rep~\cite{bhat2021deep_repara} extends it by applying reconstruction loss from image space to feature space.
BIP-Net~\cite{dudhane2022burst} obtains a restored and enhanced output image by aligning frames to use deformable convolution, exchanging information between burst features, and fusing the features progressively.
However, all the methods mentioned above do not take input bursts with non-uniform exposures for practical scenarios and assume that the first frame among the input bursts is a base frame.

\noindent \textbf{Frame selection in an image sequence.}
Existing frame selection methods mainly focus on video object segmentation and action recognition in video.
BubbleNet~\cite{griffin2019bubblenets}
chooses the best reference frame to propagate the predicted segmentation map to the other frames for video object segmentation.
ATP~\cite{buch2022revisiting} suggests using a single selected frame rather than all frames to construct video representation in action recognition. 
Dahary et al.~\cite{dahary2021digital} propose to learn exposure time in burst shots by leveraging that the long-exposed image obtains a high-SNR image and the short-exposed image is sharp.
\sh{Unlike the previous methods, we cover a scenario of input bursts with dynamic exposure times and select the base frame for each input dynamically. 

One important baseline for selecting the base frame is the Auto-Exposure (AE) algorithm.
The common approach to AE is to measure optimal exposure based on the image's histogram or entropy~\cite{zhang2006automatic,tedla2023examining}.
Zhang \textit{et al}.~\cite{zhang2006automatic} selected the best exposure time based on the highest entropy of the histogram.
Tedla \textit{et al}.~\cite{tedla2023examining} combine the content-agnostic and semantic AE algorithms. 
}

\begin{figure*}[!t]
    \centering
    \scalebox{.75}{
    \includegraphics[width=\textwidth]{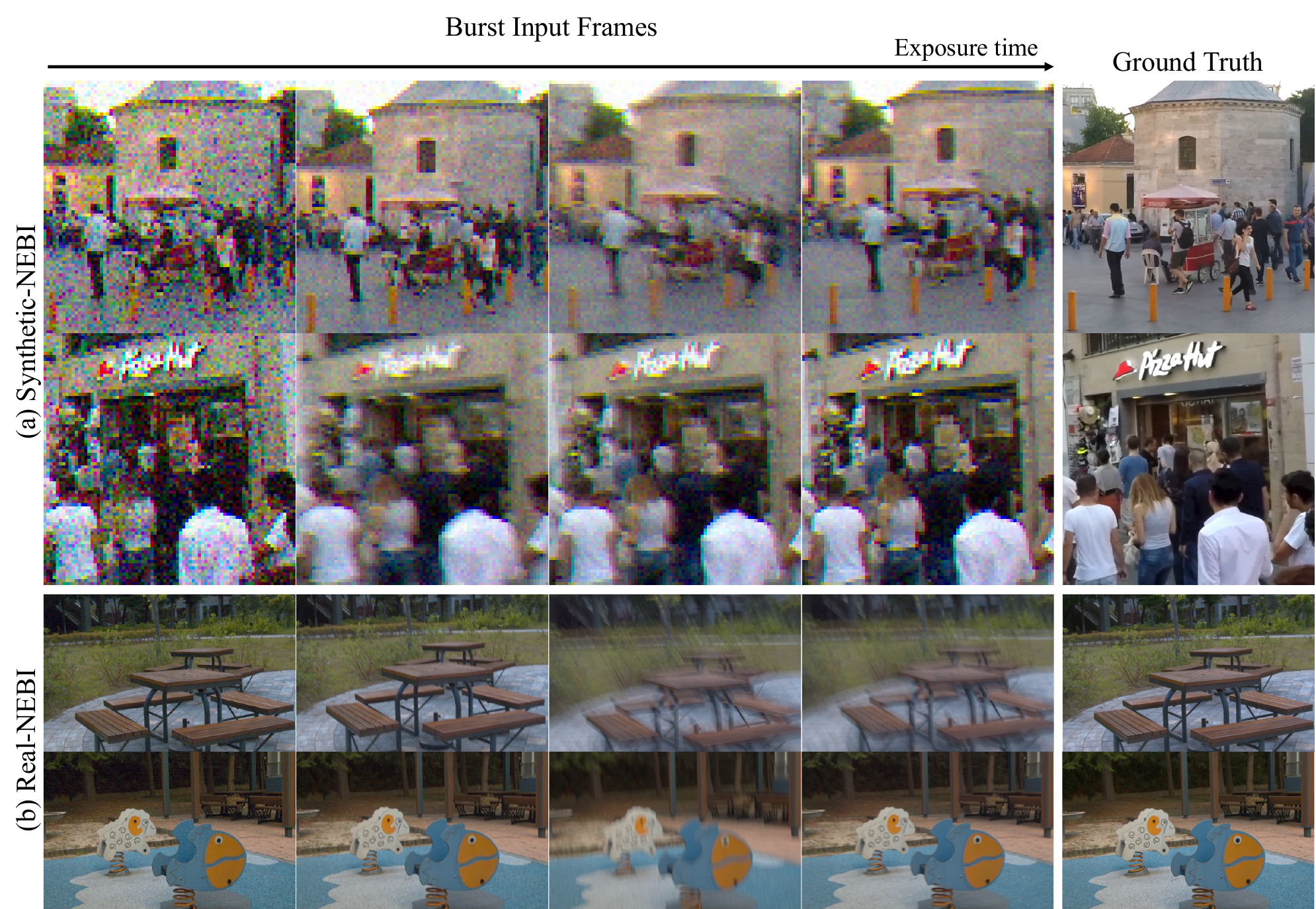}
    } \vspace{-0.3cm} 
    \caption{\textbf{Visualization of the Synthetic-/Real-NEBI dataset.} The left four columns display a subset of 14 input bursts, while the rightmost column presents the ground-truth image corresponding to the first input frame.
    From left to right in the burst sequence, the exposure time increases, resulting in decreased noise and increased blur.
    Best view to zoom.
    }
    \label{fig:nebi}
    \vspace{-0.4cm}
\end{figure*}

\section{Benchmark}
We propose a new benchmark for burst image super-resolution with non-uniform camera settings~(\textit{i.e.,} varying the exposure time), named Non-uniformly Exposed Burst Image~(NEBI). By acquiring the burst frames with various exposure times, our NEBI is able to capture a broader range of irradiance and motion characteristics within a scene for practical scenarios where the optimal exposure time is unknown.
Specifically, the benchmark consists of two tracks: Synthetic-NEBI and Real-NEBI, where Synthetic-NEBI is used for training and testing, and Real-NEBI is only used for evaluation. 
Both benchmarks have heterogeneous image noise and motion blur, and there are misalignments between bursts that naturally occur from camera movement.

\subsection{Synthetic-NEBI}
We create Synthetic-NEBI to address the cost of obtaining sufficient pairs of burst frames and the corresponding ground-truth images in the real-world. 
To achieve this, we utilize sharp frames from a video dataset~\cite{Nah_2017_CVPR} and synthesize a dataset with realistic blur and noise. 
Specifically, we synthesize blur from gyro sensor data and inject realistic noise using the Poisson-Gaussian noise model.

We simulate a longer exposure time by increasing the number of gyro sensor data and decreasing the noise amount~\cite{dahary2021digital, huang2021neighbor2neighbor, lehtinen2018noise2noise, zhang2022self}. 
This process results in a sequence of frames that resemble a burst taken from short to long exposures, allowing us to simulate non-uniform exposures and heterogeneous degradation that occur in real-world shooting environments.

\begin{figure*}[!]
   \includegraphics[width=0.90\textwidth]{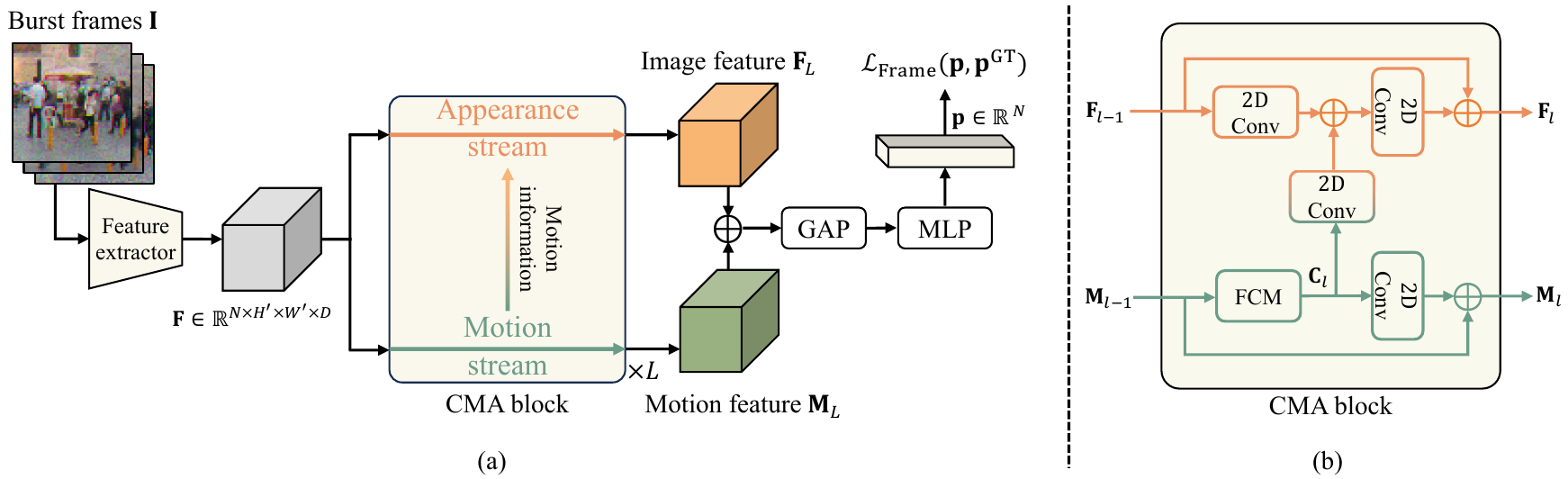}
   \vspace{-0.3cm}
   \caption{ (a)~\textbf{Overall architecture of frame selector.}
   The network begins with a feature extractor followed by $L$ Contextual Motion Aggregation (CMA) blocks, each containing appearance and motion streams that produce image feature $F_L$ and motion feature $M_L$.
   These features are summed and further refined through a Global-Averaging-Pooling (GAP) layer and an MLP layer to predict the likelihood of being the base frame.
   (b) \textbf{Feature Correlation Module~(FCM).} Given the previous motion feature $M_{l-1}$, our FCM generates the motion information $C_l$ by computing the local feature correlation. 
   The resultant $F_l$ and $M_l$ are obtained by propagating $C_l$.
   }
   \vspace{-0.3cm} 
\label{fig:overall_archi}
\end{figure*}

\wh{We simulate various degradations, including motion blur, noise, and low-resolution, considering where they occur in the image acquisition process.
During image acquisition, the scene radiance emitted from the target scene is first blurred by the relative motions between the camera and the scene, and then is blurred by the lens and sampled by the sensor. Then, it is mosaiced by the color filter array, and processed by the ISP. 
To mimic this process, we first apply the inverse ISP to obtain synthetic scene radiance, followed by applying blur synthesis, downsampling, and noise synthesis sequentially.
Specifically, }we first extract 14 consecutive images from the video dataset~\cite{Nah_2017_CVPR} and convert the images to the RAW color space through the unprocessing pipeline~\cite{brooks2019unprocessing}: inverse gamma compression, inverse gain, and inverse CCM.
We synthesize blur on the images by utilizing gyro sensor data collected from the Samsung Galaxy S22.
Specifically, considering different exposure times, we randomly sample consecutive gyro sensor values matching each burst image's exposure.
Next, we interpolate these values and then calculate homographies.
We obtain a blurred image by warping the ground truth sharp images corresponding to each burst using the previously derived homographies and subsequently averaging them.
Then, we \wh{apply downsampling and} synthesize realistic noise on the synthesized blurred images.
We randomly sample shot noise $\lambda_{shot}$ and read noise $\lambda_{read}$ parameters of the burst images according to those of exposure times.
Noisy images are synthesized using the Poisson-Gaussian noise model~\cite{brooks2019unprocessing} with the noise parameters.
\wh{Finally, we convert the burst images to the Bayer pattern RAW images through mosaicing.}
Note that the input video frames naturally contain shifting between frames for hand motion, and this is sufficient to simulate the misalignment of the synthetic burst frames.
We finally obtain a synthetic RAW burst sequence with heterogeneous image noise and motion blur.
More details for synthesizing the dataset are provided in the supplementary.

The synthetic-NEBI consists of 2,750 synthetic non-uniformly exposed RAW bursts. We split the dataset into 2,200 bursts for training and 550 bursts for the test set, using GoPro~\cite{Nah_2017_CVPR} as a source dataset.
Each burst sequence consists of 14 RAW frames center-cropped to $640 \times 640$ pixels, and the source frames are downsampled by 4 to produce $4 \times 80 \times 80$ input bursts. 
The size of the corresponding GT images is $3 \times 640 \times 640$ pixels.
The images on the left of Fig.~\ref{fig:nebi} (a) demonstrate the effect of the shorter exposure, which is sharp but noisy, while the images on the right demonstrate the effect of the longer exposure, which is high-SNR but blurry.
Overall, the Synthetic-NEBI dataset provides a comprehensive representation of non-uniform exposures shooting environments.

\subsection{Real-NEBI}
We create Real-NEBI to evaluate the quality of the result of super-resolution captured in real-world.
To collect the Real-NEBI dataset, we use a dual-camera system similar to~\cite{rim2022realistic,rim2020real, zhong2020efficient, zhong2022real}, which consists of two camera modules and a beam-splitter.
The beam-splitter equally splits photons into two camera modules; the dual-camera system can simultaneously capture burst frames with different exposures and corresponding sharp frames at the same time. 
We install a 5\% neutral density filter (ND filter) on the camera module used to capture the burst frames to ensure stable handling of long exposure time~\cite{zhong2020efficient}.
Using the camera system, we capture real-world burst frames with non-uniform exposures and the corresponding sharp and high-SNR ground-truth frame.

Specifically, one of the two camera modules captures 14 frames with gradually increasing exposure times from 0.01 to 0.14 seconds, and the other camera module captures the corresponding sharp images with 0.005 seconds of exposure time.
We inverse-proportionally decrease the gain value of each burst frame to maintain the same brightness level as the corresponding sharp frame.
We collect all images in the RAW format and perform spatial and photometric alignments, as done in~\cite{rim2022realistic}.
The burst frames are downsampled for the input of the enhancement pipeline, and the GT frames undergo only demosaicing.

The Real-NEBI dataset consists of 96 burst sequences with heterogeneous degradation by non-uniform exposure times. A burst set consists of 14 burst frames, and the input bursts and the ground-truth image have shapes of $4 \times 148 \times 238$ and $3 \times 1184 \times 1904$, respectively. Figure~\ref{fig:nebi} (b) shows the examples of our Real-NEBI dataset. 
Detailed descriptions of our Synthetic- and Real-NEBI generation are in the supplementary material.

\section{Frame Selection Network}
The goal of BISR is to reconstruct a high-quality image by merging the information from successive frames.
However, inevitable misalignments between frames occur as burst images are rapidly captured by handheld devices.
This discrepancy hampers the fusion of their complementary information, resulting in ghosting and blurring artifacts in the output image~\cite{wronski2019handheld,dudhane2022burst}.
% While existing methods primarily concentrate on developing burst alignment algorithms to address this issue, their focus lies in aligning the other frames with the first frame rather than determining which frame aligns best with the others. 
\ds{While existing methods primarily concentrate on developing burst alignment algorithms to address this issue, they focus on aligning the other frames with the first frame rather than determining which frame best aligns with the others.}
In this section, we present FSN that automatically discerns the most suitable base frame that aligns best with the others.

Given raw burst frames $\mathbf{I} \in \mathbb{R}^{N \times H \times W \times 4}$, the objective of FSN is to predict an index $k$, indicating the position of the \ds{optimal} base frame, where $N$, $H$, and $W$ represent the burst size, height, and width of a burst frame, respectively.
After the frame selection, the burst frames pass through a super-resolution network, producing an enhanced image by merging complementary information from the input burst frames into the selected base frame. 

\ds{As depicted in Figure~\ref{fig:overall_archi}, FSN first constructs the image feature of each frame using Convolutional Neural Network (CNN).}
\ds{The constructed image feature is fed into our \textbf{C}orrelation-based \textbf{M}otion \textbf{A}ware~(CMA) blocks, which update the image feature based on motion information extracted from the burst frames.}
\ds{To extract motion information from the burst frames, we propose \textbf{F}eature \textbf{C}orrelation \textbf{M}odule~(FCM), which computes the local correlation along both spatial and temporal axes.}
\ds{Finally, the output features from the last CMA block are then passed into the classifier to predict the likelihood of being the base frame.}
\ds{The remainder of this section presents the details of the feature extractor (Sec.~\ref{sec:FSN_feat}), CMA block (Sec.~\ref{sec:FSN_CMA}), and FCM (Sec.~\ref{sec:FSN_FCM}) and the training procedure (Sec.~\ref{sec:FSN_training}).}

\subsection{Feature Extractor}
\label{sec:FSN_feat}
\ds{
Given raw burst frames $\mathbf{I}$, we use two-layer CNN to extract the image feature $\mathbf{F} = [{\mathbf{f}_1;...;\mathbf{f}_N}] \in \mathbb{R}^{N \times H' \times W' \times D}$, 
where $\mathbf{f}_i \in \mathbb{R}^{H' \times W' \times D}$ is the feature of each frame and $[\cdot ; \cdot]$ is a concatenation operator. 
$H'$, $W'$, and $D$ denote the height and width of the image feature and the feature dimension, respectively.
The constructed image feature $\mathbf{F}$ is fed into the first CMA block as input.
}
\subsection{Correlation-based Motion Aware Block}
\label{sec:FSN_CMA}

To ascertain which frame aligns best with the others, the model must consider not only the degrees of degradation of each frame, as degradation contributes to misalignment issues, but also the motion information, as the alignment between frames captured by handheld devices should consider the motion trajectory.

To address this challenge, we propose a novel Correlation-based Motion Aware~(CMA) block designed to enable the model to efficiently fuse noise information within a frame and motion information across frames.
A CMA block has a two-stream architecture: an appearance stream and a motion stream.
The appearance stream updates the image feature to consider the degradation of a frame.
\ds{Simultaneously, the motion stream extracts the motion information in the frames using the Feature Correlation Module~(FCM)~(described in Sec.~\ref{sec:FSN_FCM}) and conveys motion information to the next block, independent of the image feature of the appearance stream.  }
The motion information is then propagated to the image feature, enabling the model to consider noise and motion information when selecting a base frame.
Specifically, the $l$-th CMA block takes the previous results of image feature $\mathbf{F}_{l-1}$ and motion feature $\mathbf{M}_{l-1}$, and outputs updated $\mathbf{F}_{l}$, and $\mathbf{M}_{l}$. 
In the appearance stream, the previous image feature $\mathbf{F}_{l-1}$ is updated using correlation feature $\mathbf{C}_{l}$ as follow:
\begin{align}
    g(\cdot) &= \text{ReLU}(\text{Conv2D}(\cdot)) \\
    \mathbf{F}_l &= \mathbf{F}_{l-1} + g(g(\mathbf{F}_{l-1}) + C_l), \\
    \mathbf{C}_l &= \text{FCM}(\mathbf{M}_{l-1}) \label{eq:correlation_feat}.
\end{align}
For the sake of simplicity, we omit the subscript that indicates different weights in $g$.
In equation~\ref{eq:correlation_feat}, the correlation feature $\mathbf{C}_{l}$ is extracted from the previous motion feature $\mathbf{M}_{l-1}$ using the FCM.
Similar to the appearance stream, the motion feature is also updated by using $\mathbf{C}_l$ as follows:
\begin{align}
    \mathbf{M}_l &= \mathbf{M}_{l-1} + g(\mathbf{C}_l).
\end{align}
At the first CMA block, the image feature $\mathbf{F}$ is utilized as $\mathbf{F}_0$ and $\mathbf{M}_0$.

\subsection{Feature Correlation Module}
\label{sec:FSN_FCM}
Traditional super-resolution methods typically utilize optical flow techniques to extract motion information for aligning input burst frames\cite{bhat2021deep, bhat2021deep_repara}.
However, these approaches use an off-the-shelf network pre-trained in the RGB space, resulting in inaccurate alignment when applied to RAW images, thus negatively impacting generated outputs and incurring high computational costs. 
To mitigate the issue, we utilize local feature correlation with spatial-temporal neighbors to extract motion information.

feature correlation map $\mathbf{S}\in\mathbb{R}^{N \times H' \times W' \times U \times V \times W }$ is computed as follow:
\begin{align}
    \mathbf{S}_{t,x,y,u,v,w} &= \text{sim}(\mathbf{M}_{t,x,y},\mathbf{M}_{t+u,x+v,y+w}),
\end{align}
where $\text{sim}(\cdot,\cdot)$ is similarity function (\textit{e.g.}, cosine similarity).
$U$, $V$, and $W$ are the local window sizes of the time, height, and width axes, respectively.
$(t,x,y)$ is the position of a query point and $(u,v,w)$ is a offset from the query point.
For the sake of simplicity, we omit the subscript $l$ in $\mathbf{M}$.
Then, A correlation feature $\mathbf{C}\in\mathbb{R}^{T \times X \times Y \times D}$,  which contains motion information, is extracted from $\mathbf{S}$ using a series of 3D convolution along with $U$, $V$, and $W$ axes as:
\begin{align}
    h(\cdot) &= \text{ReLU}(\text{Conv3D}(\cdot)) \\
    \mathbf{C} &= (h \circ  \cdots \circ h )(\mathbf{S}).
\end{align}
We omit the subscript that indicates different weights in $h$ for simplicity.

\subsection{Base Frame Estimation and Network Training}\label{sec:loss_functions}
\label{sec:FSN_training}

$\mathbf{F}_{L}$ and $ \mathbf{M}_{L}$, the outputs of the last CMA block, are fed into MLP after the summation to predict the probability as the base frame $\mathbf{p}\in\mathbb{R}^{N}$ as follow:
\begin{align}
    \mathbf{p} &= \sigma(\text{MLP}(\text{GAP}(\mathbf{F}_{L}+\mathbf{M}_{L}))),
\end{align}
where GAP is a global average pooling over spatial dimension and $\sigma$ is a \textit{softmax} operation.
During inference, the index with the highest probability in P is selected as a base frame. %, i.e.,:

To train our FSN,  we employ the cross-entropy between the ground-truth $\mathbf{p}^{\mathrm{GT}}$, which one-hot vector with ground-truth base frame index set to 1, and the predicted $\mathbf{p}$ as follows:
\begin{equation}
    \mathcal{L}_{\mathrm{Frame}} = \sum( -\mathbf{p}^{\textrm{GT}} * \log(\mathbf{p})  ).
\end{equation}
The ground-truth base frame index is determined by selecting the frame with the highest \sh{metric~(\textit{i.e.,} PSNR, SSIM, LPIPS)} among the input burst frames. 
\sh{Unless otherwise stated, we select the ground-truth based on the PSNR.}

\section{Experiment}
We demonstrate the effectiveness of the use of non-uniformly exposed burst shots in practical scenarios and our proposed frame selection network when integrated into existing super-resolution models~\cite{bhat2021deep, bhat2021deep_repara, dudhane2022burst}. 

\subsection{Quantitative Results}\label{sec:results_on_mcbie}
\noindent \textbf{Uniform vs.~non-uniform exposure settings.}
Table~\ref{tab:synthetic_uebi} compares PSNR scores of BIPNet~\cite{dudhane2022burst}, Deep-sr~\cite{bhat2021deep}, and Deep-rep~\cite{bhat2021deep_repara} trained on uniformly-exposed burst images at various exposure times. 
The results show that image quality increases as the model utilizes an exposure time close to the optimal.
It suggests that it is crucial to use optimally-exposed burst shots to generate high-quality images in image restoration.
However, obtaining such optimally-exposed burst shots in real-world conditions poses inherent challenges due to the dynamic nature of lighting conditions. 
Interestingly, when utilizing non-uniformly-exposed burst shots, the model exhibits comparable results to those obtained with optimal exposure times.
This demonstrates that bursts captured at non-uniform exposure settings are beneficial in practical scenarios where determining the optimal exposure time is challenging.

\begin{table}[t!]
\centering
\scalebox{0.85}{
\begin{tabular}{c|ccccc}
\toprule[1pt]
\multirow{2}{*}{Model} & \multicolumn{5}{c}{PSNR} \\ & 0.01  & 0.02  & 0.04 & 0.14 & NEBI  \\
\midrule
BIPNet~\cite{dudhane2022burst} & 33.276 & 33.311 & \textbf{34.017} & 32.749 &  \underline{33.774} \\
Deep-sr~\cite{bhat2021deep}& 33.170 & 33.288 & \textbf{33.732} & 32.845  &  \underline{33.516} \\
Deep-rep~\cite{bhat2021deep_repara}&  33.580  & 33.631   & \textbf{34.082}  &  32.918 &  \underline{34.059}\\
\bottomrule[1pt]
\end{tabular}
}
 
 % end of scale box
 \caption{\textbf{Perfomance comparison on uniform/non-uniform exposure settings.}
The table shows PSNR scores for a vanilla BISR network~\cite{bhat2021deep,bhat2021deep_repara,dudhane2022burst}, trained on the uniform exposure time from 0.01 to 0.14, and NEBI.
When the model utilizes the burst shots close to the optimal exposure time, the quality of outputs is increased.
When the optimal exposure time is unknown, it is beneficial to use the non-uniformly exposed burst shots.
 } \label{tab:synthetic_uebi}

\vspace{-0.2cm}
 
\end{table}

\begin{table}[t!]

\centering
\scalebox{0.70}{
\begin{tabular}{c|cc|ccc}
\toprule[1pt]
Dataset  & Model  & Frame selection & PSNR$\uparrow$  & SSIM$\uparrow$ & LPIPS$\downarrow$\\ \midrule
\multirow{6}{*}{Synthetic-NEBI}

& \multirow{2}{*}{BIPNet~\cite{dudhane2022burst}} & - & 33.774 & 0.920 & 0.108       \\
& & FSN & \textbf{33.878} & \textbf{0.922} & \textbf{0.106}    \\
\cmidrule{2-6}

& \multirow{2}{*}{Deep-sr~\cite{bhat2021deep}}  
& - & 33.516 & 0.917 & 0.117      \\                    
& & FSN & \textbf{33.872} & \textbf{0.921} &  \textbf{0.107} \\ 
\cmidrule{2-6}

& \multirow{2}{*}{Deep-rep~\cite{bhat2021deep_repara}}             
& - &34.059 & 0.925 & 0.103      \\
& & FSN & \textbf{34.313}   & \textbf{0.928}   & \textbf{0.099} \\
\midrule

\multirow{9}{*}{Real-NEBI}
& \multirow{3}{*}{BIPNet~\cite{dudhane2022burst}}
& - & 33.957 & \textbf{0.904} & \textbf{0.143}       \\
& & AE & 34.184 & 0.900 & 0.157 \\
& & FSN & \textbf{34.367} & 0.901 & 0.157    \\
\cmidrule{2-6}

& \multirow{3}{*}{Deep-sr~\cite{bhat2021deep}}  
& - & 31.084 & \textbf{0.908} & 0.180      \\
& & AE & 30.914 & 0.901 & 0.195 \\
& & FSN & \textbf{31.149} & 0.907 & \textbf{0.180}    \\
\cmidrule{2-6}

& \multirow{3}{*}{Deep-rep~\cite{bhat2021deep_repara}}             
& -  & 31.088 & \textbf{0.911} & \textbf{0.174}      \\
& & AE & 30.861 & 0.903 & 0.193 \\
& & FSN & \textbf{31.308} & 0.904 & 0.194    \\

\bottomrule
\end{tabular}
}
 % end of scale box
 \caption{\textbf{Results on Synthetic-/Real- NEBI.}
`AE' and 'FSN' stand for Auto Exposure and Frame Selection Network, respectively. The best score is highlighted in bold. %This result shows that our FSN improves the qualities of super-resolution results.
 } \label{tab:synthetic_meb_results_scratch}
 \vspace{-0.5cm}
\end{table}

\noindent \textbf{Effect of FSN on Synthetic-/Real-NEBI.}
Table~\ref{tab:synthetic_meb_results_scratch} shows a quantitative comparison between existing super-resolution models and their variants with our FSN on Synthetic-/Real-NEBI. 
On the Synthetic-NEBI, the existing models exhibit subpar performance due to designating the first frame as the base frame.
Compared to existing models, their variants incorporating our FSN consistently demonstrate performance improvements.
These results indicate that selecting an appropriate base frame contributes to the performance increase, facilitating improved alignment and fusion among the burst frames.
For the Real-NEBI, we evaluate the performance of our FSN on Real-NEBI while training on Synthetic-NEBI as acquiring labeled data for super-resolution is challenging and costly.
Similar to Snythetic-NEBI, Our FSN consistently improves the PSNR values of existing super-resolution models on the real-world burst examples of non-uniform exposure conditions, despite being trained on Synthetic-NEBI.
These results show the importance of frame selection even in real-world scenarios, demonstrating the effectiveness of our FSN.
Our FSN also shows comparable results on SSIM and LPIPS metrics.
% These results are attributed to training the FSN with the frame that yields the highest PSNR.

\sh{\noindent{\textbf{Auto-Exposure~(AE) vs. FSN.}}
To demonstrate the effectiveness of our FSN, we also compare FSN with the AE algorithm. Following previous AE methods~\cite{tedla2023examining,zhang2006automatic}, we select the base frame with the maximum entropy calculated based on the image's histogram.
In Table~\ref{tab:synthetic_meb_results_scratch}, we observe marginal PSNR improvement or even declines when using AE.
This suggests that AE is sub-optimal in selecting the base frame, emphasizing the effectiveness of FSN.
}

\sh{\noindent{\textbf{Target frames for FSN training.}}
As we train the FSN to select the frame that yields the highest PSNR on Synthetic-NEBI, the FSN may exhibit sub-optimal results on other metrics, such as SSIM and LPIPS, when applied to Real-NEBI.
In Table~\ref{tab:frame_selection_methods}, we explore using different target frames in training FSN and evaluate its effect on performance in Real-NEBI.
Specifically, we trained FSN to select the frame that yields the highest PSNR, SSIM, and LPIPS on the Synthetic-NEBI and evaluate the FSN on Real-NEBI.
The results show that training based on SSIM yields high SSIM performance, while training based on LPIPS yields high LPIPS performance.
}

\subsection{Qualitative results}\label{sec:qualitative}
\begin{figure*}[t!]
    \centering
    \scalebox{0.87}{
    \includegraphics[width=2\columnwidth]{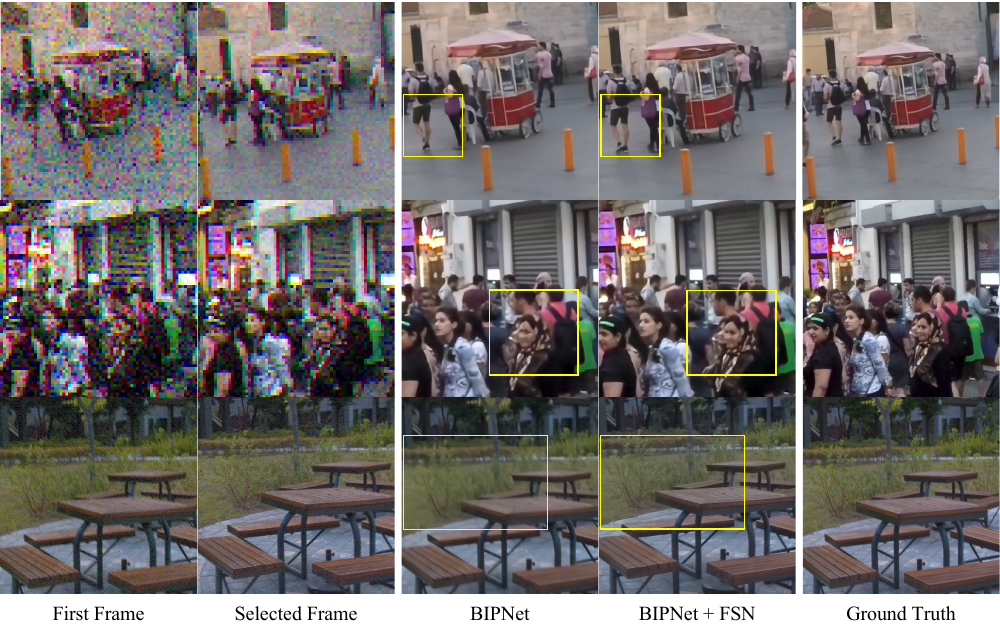}
    } 
    \vspace{-0.4cm} 
    \caption{\textbf{Qualitative comparison on synthetic-NEBI (top two rows) and real-NEBI (bottom row).} BIPNet~\cite{dudhane2022burst} with our frame selector enhances high-frequency image details by merging effectively complementary information from multiple frames into the selected base frame, while BIPNet~\cite{dudhane2022burst} predicts blurry images due to the severe degradations in the first frame.    
    Best view to zoom.}
    \vspace{-0.5cm}
    \label{fig:qualitative_mcbie_burstsr}
\end{figure*}

We visualize super-resolution results and the selected base frame to verify the effectiveness of our FSN in Figure~\ref{fig:qualitative_mcbie_burstsr}.
The first and second columns show the first frames, which are typically used as a base frame, and frames selected as base frames by FSN, respectively.
We observe that our proposed method selects frames with less degradation.
Additionally, we note increased qualities of super-resolution results when applying FSN to Deep-rep (third vs fourth columns).
These results indicate that selecting an appropriate frame for the base frame increases the image quality in the super-resolution task. 
For example, when the base frame is noisy, the legs appear blurry in the first row. However, when selecting a frame with less degradation, the legs appear sharp compared to the Deep-rep~\cite{bhat2021deep_repara}.

\subsection{Ablation Study of FSN}\label{sec:ablation_study}
We explore the impact of each individual component of FSN.
For the ablation study of FSN, we adopt Deep-rep~\cite{bhat2021deep_repara} trained/tested on Synthetic-NEBI as a super-resolution network. Specifically, we attach variants of FSN to Deep-rep and evaluate their performance.

\noindent{\textbf{Number of CMA blocks.}}
We utilize the CMA block to encode visual features with motion information.
To investigate the impact of the number of CMA blocks, we gradually increase the number of CMA blocks to a vanilla FSN that has CMA blocks.
Comparing the baseline (solo super-resolution network) and the baseline variants with the vanilla FSN, there is notable performance gain as shown in the first and second rows of Table~\ref{tab:ablation_study_cma_block}. This result demonstrates that selecting an appropriate base frame contributes to the model's ability to generate high-quality images.
Furthermore, the addition of CMA blocks increases the performance gap between the baseline and the baseline with the FSN.
we observed that employing two CMA blocks yields the best performance across all metrics.

We also increase the capacity of the vanilla FSN to have a similar number of parameters with the FSN with 2 CMA blocks (\textit{i.e.}, $0^{*}$) to investigate whether the improvement is due to an increased number of parameters or the effectiveness of our CMA block.
The results show that even though the capacity of the vanilla FSN is increased, the FSN with the CMA blocks outperforms the vanilla FSN.
These results demonstrate the effectiveness of our CMA block.

\begin{table}[t!]

\centering
\scalebox{0.85}{
\begin{tabular}{c|ccc}
\toprule[1pt] 
Frame selection  & PSNR$\uparrow$ & SSIM$\uparrow$ & LPIPS$\downarrow$ \\
\midrule
- & 33.957 & 0.904 & 0.143 \\
\textrm{FSN}$_{P}$ & 34.367 & 0.901 & 0.157  \\
\textrm{FSN}$_{S}$ &  \textbf{34.675} & \textbf{0.907} & 0.155   \\
\textrm{FSN}$_{L}$ &  34.053 & 0.905 & \textbf{0.143} \\
\textrm{oracle} & 35.650 & 0.912 & 0.142 \\ 
\bottomrule
\end{tabular}
}
 \caption{\textbf{Comparison of training strategies on Real-NEBI.}
The table illustrates how performance varies when different metrics are employed to select the ground-truth frame in the Real-NEBI dataset
The subscription $P,S,L$ denote frame selection strategies based on PSNR, SSIM, and LPIPS scores respectively.
 }
\label{tab:frame_selection_methods}
 \vspace{-0.2cm}
\end{table}

\begin{table}[t!]

\centering
\scalebox{0.85}{

\begin{tabular}{cccc}
\toprule[1pt]
\# of CMA block & PSNR$\uparrow$ & SSIM$\uparrow$ & LPIPS$\downarrow$  \\
\midrule
- & 34.059 & 0.925 & 0.103 \\ 
\midrule
0 & 34.174 &  0.926 & 0.101 \\
0$^{*}$ & 34.164 &  0.926 & 0.101 \\
1 & 34.252 &  0.927 & 0.101 \\
2 & \textbf{34.313} & \textbf{0.928} & \textbf{0.099} \\
4 & 34.248 & 0.927 & 0.100 \\

\bottomrule[1pt]
\end{tabular}
}
\caption{\textbf{Ablation study on the number of CMA Blocks.}
The first row represents the baseline, \textit{i.e.}, vanilla super-resolution network, while the subsequent five rows explore the impact of varying numbers of CMA blocks. * denotes the balanced model size as the number of parameters when there are two of the CMA block. 
 } \label{tab:ablation_study_cma_block}
\vspace{-0.2cm}
\end{table}

\begin{table}[t!]
\centering
\scalebox{0.80}{
\begin{tabular}{ccc|ccc}
\toprule[1pt]
Appear.  & Motion & Motion Feat & PSNR$\uparrow$ & SSIM$\uparrow$ & LPIPS$\downarrow$  \\
\midrule
- & - & - & 34.174 & 0.926 & 0.101\\
\checkmark & - & - &34.123 & 0.926 & 0.103\\
 -  & \checkmark & PWCNet$_{\textrm{raw}}$ &  34.104 & 0.926 & 0.102\\
 - & \checkmark & PWCNet$_{\textrm{rgb}}$ &  34.088 & 0.925 & 0.102 \\
  - & \checkmark & FCM & 34.266 &  0.927 & 0.101\\
\checkmark & \checkmark & FCM &\textbf{34.313} & \textbf{0.928} & \textbf{0.099} \\
\bottomrule[1pt]
\end{tabular}
}
\caption{\textbf{Impact of appearance \& motion streams.}
We compare the performance of our frame selector with different combinations of appearance and motion streams. The 'Motion Feat' column indicates the ways of extracting the motion feature.
 } \label{tab:ablation_study_appearance_motion_stream}
\vspace{-0.5cm}
\end{table}

\noindent{\textbf{Appearance \& motion streams.}}
Our CMA block consists of the appearance stream to update the image feature and the motion stream to encode the motion information in the input burst frames.
To investigate the impact of each stream, we gradually add each stream to the baseline, where neither appearance nor motion stream is applied~(\textit{i.e.}, vanilla FSN) in Table~\ref{tab:ablation_study_appearance_motion_stream}.
The appearance stream in isolation fails to demonstrate discernible improvement compared to the baseline \sh{as there is a misalignment between the burst shots.}
However, the contribution of the motion stream is significant, indicating that motion information is crucial to the base frame selection.
Notably, leveraging both appearance and motion streams provides the best performance as the image feature in the appearance stream is updated by using the correct sub-pixel information aligned by the FCM.

\sh{
\noindent{\textbf{Impact of FCM.}}
In Table~\ref{tab:ablation_study_appearance_motion_stream}, we replace the FCM with an off-the-shelf optical flow network, PWCNet, to investigate the effectiveness of our FCM.
Following Deep-sr~\cite{bhat2021deep}, we utilize the raw image to obtain the optical flow, PWCNet$_{\textrm{raw}}$.
The result shows that utilizing optical flow for motion information extraction leads to a decrease in the model's performance compared to utilizing our FCM.
Since the optical flow network is trained on RGB, it may result in inaccurate predictions on raw images, negatively impacting the model's performance.
Therefore, we also utilize the RGB image to obtain the optical flow, PWCNet$_{\textrm{rgb}}$.
Interestingly, the model's performance is degraded compared to PWCNet$_{\textrm{raw}}$.
These results indicate that optical flow may not be suitable for extracting motion features for frame selection, highlighting the effectiveness of FCM.
}

\begin{table}[t!]
\centering
\scalebox{0.85}{
\begin{tabular}{c|cc}
\toprule[1pt]
Model &Params. & FLOPs \\
\midrule
BIPNet ~\cite{dudhane2022burst} & 6.67M & 1.67T\\
Deep-sr ~\cite{bhat2021deep} & 13.01M & 544.86G \\
Deep-rep ~\cite{bhat2021deep_repara} & 12.13M & 1.28T \\
\midrule
FSN & 1.59M & 18.53G \\
\bottomrule
\end{tabular}
}
\caption{\textbf{Parameter comparison between existing methods and FSN.}
We analyze the number of parameters and FLOPs of FSN in contrast to previous burst enhancement models.
 } \label{tab:ablation_study_parameter}
 % end of scale box
\vspace{-0.6cm}
\end{table}

\noindent{\textbf{Parameter Analysis.}
Applying our FSN to existing super-resolution networks may increase computational costs, potentially impacting performance metrics such as FPS and memory usage.
We conduct an analysis of the number of parameters and FLOPs in Table~\ref{tab:ablation_study_parameter}.
FSN has a compact model size and small FLOPs compared to existing burst enhancement methods ~\cite{bhat2021deep, bhat2021deep_repara, dudhane2022burst}.
This guarantees plug-and-play integration of FSN into existing models with a low computational overhead.

\section{Conclusion}
We have explored using burst shots with non-uniform exposures for practical scenarios where the acquisition of an optimal exposure time is challenging.
By using burst shots with non-uniform exposures, the model reconstructs a high-resolution image. 
As burst shots with non-uniform exposures present various degrees of degradation, we introduce a base frame selection model to improve the performance. 
Our evaluations on the Synthetic- and Real-NEBI datasets demonstrated the effectiveness of using non-uniformly exposed burst shots for practical scenarios and the importance of selecting a proper base frame. 
We believe that our proposed dataset and method can contribute to the advancement of super-resolution methods and pave the way for further research in this field.

\noindent \textbf{Acknowledgements.}
This work was supported by Samsung Advanced Institute of Technology (SAIT), the NRF grant (No. 2023R1A2C200494611), and the IITP grants (No. 2022-0-00290, Visual Intelligence for Space-Time Understanding and Generation, No.2019-0-01906, AI Graduate School Program at POSTECH) funded by the Korea government.

{
    \small
    \bibliographystyle{ieeenat_fullname}
    \bibliography{main}

\begin{thebibliography}{50}
\providecommand{\natexlab}[1]{#1}
\providecommand{\url}[1]{\texttt{#1}}
\expandafter\ifx\csname urlstyle\endcsname\relax
  \providecommand{\doi}[1]{doi: #1}\else
  \providecommand{\doi}{doi: \begingroup \urlstyle{rm}\Url}\fi

\bibitem[Aittala and Durand(2018)]{aittala2018burst}
Miika Aittala and Fr{\'e}do Durand.
\newblock Burst image deblurring using permutation invariant convolutional
  neural networks.
\newblock In \emph{Proceedings of the European conference on computer vision
  (ECCV)}, pages 731--747, 2018.

\bibitem[Bhat et~al.(2021{\natexlab{a}})Bhat, Danelljan, Van~Gool, and
  Timofte]{bhat2021deep}
Goutam Bhat, Martin Danelljan, Luc Van~Gool, and Radu Timofte.
\newblock Deep burst super-resolution.
\newblock In \emph{Proceedings of the IEEE/CVF Conference on Computer Vision
  and Pattern Recognition}, pages 9209--9218, 2021{\natexlab{a}}.

\bibitem[Bhat et~al.(2021{\natexlab{b}})Bhat, Danelljan, Yu, Van~Gool, and
  Timofte]{bhat2021deep_repara}
Goutam Bhat, Martin Danelljan, Fisher Yu, Luc Van~Gool, and Radu Timofte.
\newblock Deep reparametrization of multi-frame super-resolution and denoising.
\newblock In \emph{Proceedings of the IEEE/CVF International Conference on
  Computer Vision}, pages 2460--2470, 2021{\natexlab{b}}.

\bibitem[Brooks et~al.(2019)Brooks, Mildenhall, Xue, Chen, Sharlet, and
  Barron]{brooks2019unprocessing}
Tim Brooks, Ben Mildenhall, Tianfan Xue, Jiawen Chen, Dillon Sharlet, and
  Jonathan~T Barron.
\newblock Unprocessing images for learned raw denoising.
\newblock In \emph{Proceedings of the IEEE/CVF Conference on Computer Vision
  and Pattern Recognition}, pages 11036--11045, 2019.

\bibitem[Buch et~al.(2022)Buch, Eyzaguirre, Gaidon, Wu, Fei-Fei, and
  Niebles]{buch2022revisiting}
Shyamal Buch, Crist{\'o}bal Eyzaguirre, Adrien Gaidon, Jiajun Wu, Li Fei-Fei,
  and Juan~Carlos Niebles.
\newblock Revisiting the" video" in video-language understanding.
\newblock In \emph{Proceedings of the IEEE/CVF Conference on Computer Vision
  and Pattern Recognition}, pages 2917--2927, 2022.

\bibitem[Chen et~al.(2018)Chen, Chen, Xu, and Koltun]{chen2018learning}
Chen Chen, Qifeng Chen, Jia Xu, and Vladlen Koltun.
\newblock Learning to see in the dark.
\newblock In \emph{Proceedings of the IEEE conference on computer vision and
  pattern recognition}, pages 3291--3300, 2018.

\bibitem[Cornebise et~al.(2022)Cornebise, Or{\v{s}}oli{\'c}, and
  Kalaitzis]{cornebise2022open}
Julien Cornebise, Ivan Or{\v{s}}oli{\'c}, and Freddie Kalaitzis.
\newblock Open high-resolution satellite imagery: The worldstrat dataset--with
  application to super-resolution.
\newblock \emph{Advances in Neural Information Processing Systems},
  35:\penalty0 25979--25991, 2022.

\bibitem[Dahary et~al.(2021)Dahary, Jacoby, and Bronstein]{dahary2021digital}
Omer Dahary, Matan Jacoby, and Alex~M Bronstein.
\newblock Digital gimbal: End-to-end deep image stabilization with learnable
  exposure times.
\newblock In \emph{Proceedings of the IEEE/CVF Conference on Computer Vision
  and Pattern Recognition}, pages 11936--11945, 2021.

\bibitem[Deudon et~al.(2020)Deudon, Kalaitzis, Goytom, Arefin, Lin, Sankaran,
  Michalski, Kahou, Cornebise, and Bengio]{deudon2020highres}
Michel Deudon, Alfredo Kalaitzis, Israel Goytom, Md~Rifat Arefin, Zhichao Lin,
  Kris Sankaran, Vincent Michalski, Samira~E Kahou, Julien Cornebise, and
  Yoshua Bengio.
\newblock Highres-net: Recursive fusion for multi-frame super-resolution of
  satellite imagery.
\newblock \emph{arXiv preprint arXiv:2002.06460}, 2020.

\bibitem[Dong et~al.(2015)Dong, Loy, He, and Tang]{dong2015image}
Chao Dong, Chen~Change Loy, Kaiming He, and Xiaoou Tang.
\newblock Image super-resolution using deep convolutional networks.
\newblock \emph{IEEE transactions on pattern analysis and machine
  intelligence}, 38\penalty0 (2):\penalty0 295--307, 2015.

\bibitem[Dudhane et~al.(2022)Dudhane, Zamir, Khan, Khan, and
  Yang]{dudhane2022burst}
Akshay Dudhane, Syed~Waqas Zamir, Salman Khan, Fahad~Shahbaz Khan, and
  Ming-Hsuan Yang.
\newblock Burst image restoration and enhancement.
\newblock In \emph{Proceedings of the IEEE/CVF Conference on Computer Vision
  and Pattern Recognition}, pages 5759--5768, 2022.

\bibitem[Elad and Feuer(1997)]{elad1997restoration}
Michael Elad and Arie Feuer.
\newblock Restoration of a single superresolution image from several blurred,
  noisy, and undersampled measured images.
\newblock \emph{IEEE transactions on image processing}, 6\penalty0
  (12):\penalty0 1646--1658, 1997.

\bibitem[Farsiu et~al.(2004)Farsiu, Robinson, Elad, and
  Milanfar]{farsiu2004fast}
Sina Farsiu, M~Dirk Robinson, Michael Elad, and Peyman Milanfar.
\newblock Fast and robust multiframe super resolution.
\newblock \emph{IEEE transactions on image processing}, 13\penalty0
  (10):\penalty0 1327--1344, 2004.

\bibitem[Godard et~al.(2018)Godard, Matzen, and Uyttendaele]{godard2018deep}
Cl{\'e}ment Godard, Kevin Matzen, and Matt Uyttendaele.
\newblock Deep burst denoising.
\newblock In \emph{Proceedings of the European conference on computer vision
  (ECCV)}, pages 538--554, 2018.

\bibitem[Greenspan(2009)]{greenspan2009super}
Hayit Greenspan.
\newblock Super-resolution in medical imaging.
\newblock \emph{The computer journal}, 52\penalty0 (1):\penalty0 43--63, 2009.

\bibitem[Griffin and Corso(2019)]{griffin2019bubblenets}
Brent~A Griffin and Jason~J Corso.
\newblock Bubblenets: Learning to select the guidance frame in video object
  segmentation by deep sorting frames.
\newblock In \emph{Proceedings of the IEEE/CVF Conference on Computer Vision
  and Pattern Recognition}, pages 8914--8923, 2019.

\bibitem[Hardie(2007)]{hardie2007fast}
Russell Hardie.
\newblock A fast image super-resolution algorithm using an adaptive wiener
  filter.
\newblock \emph{IEEE Transactions on Image Processing}, 16\penalty0
  (12):\penalty0 2953--2964, 2007.

\bibitem[Haris et~al.(2021)Haris, Shakhnarovich, and Ukita]{haris2021task}
Muhammad Haris, Greg Shakhnarovich, and Norimichi Ukita.
\newblock Task-driven super resolution: Object detection in low-resolution
  images.
\newblock In \emph{Neural Information Processing: 28th International
  Conference, ICONIP 2021, Sanur, Bali, Indonesia, December 8--12, 2021,
  Proceedings, Part V 28}, pages 387--395. Springer, 2021.

\bibitem[Hasinoff et~al.(2016)Hasinoff, Sharlet, Geiss, Adams, Barron, Kainz,
  Chen, and Levoy]{hasinoff2016burst}
Samuel~W Hasinoff, Dillon Sharlet, Ryan Geiss, Andrew Adams, Jonathan~T Barron,
  Florian Kainz, Jiawen Chen, and Marc Levoy.
\newblock Burst photography for high dynamic range and low-light imaging on
  mobile cameras.
\newblock \emph{ACM Transactions on Graphics (ToG)}, 35\penalty0 (6):\penalty0
  1--12, 2016.

\bibitem[Huang et~al.(2021)Huang, Li, Jia, Lu, and
  Liu]{huang2021neighbor2neighbor}
Tao Huang, Songjiang Li, Xu Jia, Huchuan Lu, and Jianzhuang Liu.
\newblock Neighbor2neighbor: Self-supervised denoising from single noisy
  images.
\newblock In \emph{Proceedings of the IEEE/CVF conference on computer vision
  and pattern recognition}, pages 14781--14790, 2021.

\bibitem[Isaac and Kulkarni(2015)]{isaac2015super}
Jithin~Saji Isaac and Ramesh Kulkarni.
\newblock Super resolution techniques for medical image processing.
\newblock In \emph{2015 International Conference on Technologies for
  Sustainable Development (ICTSD)}, pages 1--6. IEEE, 2015.

\bibitem[Karadeniz et~al.(2021)Karadeniz, Erdem, and Erdem]{karadeniz2021burst}
Ahmet~Serdar Karadeniz, Erkut Erdem, and Aykut Erdem.
\newblock Burst photography for learning to enhance extremely dark images.
\newblock \emph{IEEE Transactions on Image Processing}, 30:\penalty0
  9372--9385, 2021.

\bibitem[K{\"o}hler(2018)]{kohler2018multi}
Thomas K{\"o}hler.
\newblock Multi-frame super-resolution reconstruction with applications to
  medical imaging.
\newblock \emph{arXiv preprint arXiv:1812.09375}, 2018.

\bibitem[Lecouat et~al.(2021)Lecouat, Ponce, and Mairal]{lecouat2021lucas}
Bruno Lecouat, Jean Ponce, and Julien Mairal.
\newblock Lucas-kanade reloaded: End-to-end super-resolution from raw image
  bursts.
\newblock In \emph{Proceedings of the IEEE/CVF International Conference on
  Computer Vision}, pages 2370--2379, 2021.

\bibitem[Lecouat et~al.(2022)Lecouat, Eboli, Ponce, and
  Mairal]{lecouat2022high}
Bruno Lecouat, Thomas Eboli, Jean Ponce, and Julien Mairal.
\newblock High dynamic range and super-resolution from raw image bursts.
\newblock \emph{arXiv preprint arXiv:2207.14671}, 2022.

\bibitem[Lehtinen et~al.(2018)Lehtinen, Munkberg, Hasselgren, Laine, Karras,
  Aittala, and Aila]{lehtinen2018noise2noise}
Jaakko Lehtinen, Jacob Munkberg, Jon Hasselgren, Samuli Laine, Tero Karras,
  Miika Aittala, and Timo Aila.
\newblock Noise2noise: Learning image restoration without clean data.
\newblock In \emph{International Conference on Machine Learning}, pages
  2965--2974. PMLR, 2018.

\bibitem[Liba et~al.(2019)Liba, Murthy, Tsai, Brooks, Xue, Karnad, He, Barron,
  Sharlet, Geiss, et~al.]{liba2019handheld}
Orly Liba, Kiran Murthy, Yun-Ta Tsai, Tim Brooks, Tianfan Xue, Nikhil Karnad,
  Qiurui He, Jonathan~T Barron, Dillon Sharlet, Ryan Geiss, et~al.
\newblock Handheld mobile photography in very low light.
\newblock \emph{ACM Trans. Graph.}, 38\penalty0 (6):\penalty0 164--1, 2019.

\bibitem[Liu et~al.(2014)Liu, Yuan, Tang, Uyttendaele, and Sun]{liu2014fast}
Ziwei Liu, Lu Yuan, Xiaoou Tang, Matt Uyttendaele, and Jian Sun.
\newblock Fast burst images denoising.
\newblock \emph{ACM Transactions on Graphics (TOG)}, 33\penalty0 (6):\penalty0
  1--9, 2014.

\bibitem[Mahapatra et~al.(2019)Mahapatra, Bozorgtabar, and
  Garnavi]{mahapatra2019image}
Dwarikanath Mahapatra, Behzad Bozorgtabar, and Rahil Garnavi.
\newblock Image super-resolution using progressive generative adversarial
  networks for medical image analysis.
\newblock \emph{Computerized Medical Imaging and Graphics}, 71:\penalty0
  30--39, 2019.

\bibitem[Musunuri et~al.(2022)Musunuri, Kwon, and Kung]{musunuri2022srodnet}
Yogendra~Rao Musunuri, Oh-Seol Kwon, and Sun-Yuan Kung.
\newblock Srodnet: Object detection network based on super resolution for
  autonomous vehicles.
\newblock \emph{Remote Sensing}, 14\penalty0 (24):\penalty0 6270, 2022.

\bibitem[Nah et~al.(2017)Nah, Kim, and Lee]{Nah_2017_CVPR}
Seungjun Nah, Tae~Hyun Kim, and Kyoung~Mu Lee.
\newblock Deep multi-scale convolutional neural network for dynamic scene
  deblurring.
\newblock In \emph{CVPR}, 2017.

\bibitem[Nichol et~al.(2021)Nichol, Dhariwal, Ramesh, Shyam, Mishkin, McGrew,
  Sutskever, and Chen]{nichol2021glide}
Alex Nichol, Prafulla Dhariwal, Aditya Ramesh, Pranav Shyam, Pamela Mishkin,
  Bob McGrew, Ilya Sutskever, and Mark Chen.
\newblock Glide: Towards photorealistic image generation and editing with
  text-guided diffusion models.
\newblock \emph{arXiv preprint arXiv:2112.10741}, 2021.

\bibitem[Noh et~al.(2019)Noh, Bae, Lee, Seo, and Kim]{noh2019better}
Junhyug Noh, Wonho Bae, Wonhee Lee, Jinhwan Seo, and Gunhee Kim.
\newblock Better to follow, follow to be better: Towards precise supervision of
  feature super-resolution for small object detection.
\newblock In \emph{Proceedings of the IEEE/CVF International Conference on
  Computer Vision}, pages 9725--9734, 2019.

\bibitem[Ramesh et~al.(2022)Ramesh, Dhariwal, Nichol, Chu, and
  Chen]{ramesh2022hierarchical}
Aditya Ramesh, Prafulla Dhariwal, Alex Nichol, Casey Chu, and Mark Chen.
\newblock Hierarchical text-conditional image generation with clip latents.
  arxiv 2022.
\newblock \emph{arXiv preprint arXiv:2204.06125}, 2022.

\bibitem[Rim et~al.(2020)Rim, Lee, Won, and Cho]{rim2020real}
Jaesung Rim, Haeyun Lee, Jucheol Won, and Sunghyun Cho.
\newblock Real-world blur dataset for learning and benchmarking deblurring
  algorithms.
\newblock In \emph{Computer Vision--ECCV 2020: 16th European Conference,
  Glasgow, UK, August 23--28, 2020, Proceedings, Part XXV 16}, pages 184--201.
  Springer, 2020.

\bibitem[Rim et~al.(2022)Rim, Kim, Kim, Lee, Lee, and Cho]{rim2022realistic}
Jaesung Rim, Geonung Kim, Jungeon Kim, Junyong Lee, Seungyong Lee, and Sunghyun
  Cho.
\newblock Realistic blur synthesis for learning image deblurring.
\newblock \emph{arXiv preprint arXiv:2202.08771}, 2022.

\bibitem[Rombach et~al.(2022)Rombach, Blattmann, Lorenz, Esser, and
  Ommer]{rombach2022high}
Robin Rombach, Andreas Blattmann, Dominik Lorenz, Patrick Esser, and Bj{\"o}rn
  Ommer.
\newblock High-resolution image synthesis with latent diffusion models.
\newblock In \emph{Proceedings of the IEEE/CVF conference on computer vision
  and pattern recognition}, pages 10684--10695, 2022.

\bibitem[Shermeyer and Van~Etten()]{shermeyer1812effects}
J Shermeyer and A Van~Etten.
\newblock The effects of super-resolution on object detection performance in
  satellite imagery. arxiv 2019.
\newblock \emph{arXiv preprint arXiv:1812.04098}.

\bibitem[Takeda et~al.(2007)Takeda, Farsiu, and Milanfar]{takeda2007kernel}
Hiroyuki Takeda, Sina Farsiu, and Peyman Milanfar.
\newblock Kernel regression for image processing and reconstruction.
\newblock \emph{IEEE Transactions on image processing}, 16\penalty0
  (2):\penalty0 349--366, 2007.

\bibitem[Tedla et~al.(2023)Tedla, Yang, and Brown]{tedla2023examining}
SaiKiran Tedla, Beixuan Yang, and Michael~S Brown.
\newblock Examining autoexposure for challenging scenes.
\newblock In \emph{Proceedings of the IEEE/CVF International Conference on
  Computer Vision}, pages 13076--13085, 2023.

\bibitem[Tsai(1984)]{tsai1984multiframe}
R Tsai.
\newblock Multiframe image restoration and registration.
\newblock \emph{Advance Computer Visual and Image Processing}, 1:\penalty0
  317--339, 1984.

\bibitem[Wieschollek et~al.(2017)Wieschollek, Sch{\"o}lkopf, Lensch, and
  Hirsch]{wieschollek2017end}
Patrick Wieschollek, Bernhard Sch{\"o}lkopf, Hendrik~PA Lensch, and Michael
  Hirsch.
\newblock End-to-end learning for image burst deblurring.
\newblock In \emph{Computer Vision--ACCV 2016: 13th Asian Conference on
  Computer Vision, Taipei, Taiwan, November 20-24, 2016, Revised Selected
  Papers, Part IV 13}, pages 35--51. Springer, 2017.

\bibitem[Wronski et~al.(2019)Wronski, Garcia-Dorado, Ernst, Kelly, Krainin,
  Liang, Levoy, and Milanfar]{wronski2019handheld}
Bartlomiej Wronski, Ignacio Garcia-Dorado, Manfred Ernst, Damien Kelly, Michael
  Krainin, Chia-Kai Liang, Marc Levoy, and Peyman Milanfar.
\newblock Handheld multi-frame super-resolution.
\newblock \emph{ACM Transactions on Graphics (TOG)}, 38\penalty0 (4):\penalty0
  1--18, 2019.

\bibitem[Zamir et~al.(2020)Zamir, Arora, Khan, Hayat, Khan, Yang, and
  Shao]{zamir2020learning}
Syed~Waqas Zamir, Aditya Arora, Salman Khan, Munawar Hayat, Fahad~Shahbaz Khan,
  Ming-Hsuan Yang, and Ling Shao.
\newblock Learning enriched features for real image restoration and
  enhancement.
\newblock In \emph{European Conference on Computer Vision}, pages 492--511.
  Springer, 2020.

\bibitem[Zamir et~al.(2021)Zamir, Arora, Khan, Hayat, Khan, Yang, and
  Shao]{zamir2021multi}
Syed~Waqas Zamir, Aditya Arora, Salman Khan, Munawar Hayat, Fahad~Shahbaz Khan,
  Ming-Hsuan Yang, and Ling Shao.
\newblock Multi-stage progressive image restoration.
\newblock In \emph{Proceedings of the IEEE/CVF conference on computer vision
  and pattern recognition}, pages 14821--14831, 2021.

\bibitem[Zhang et~al.(2006)Zhang, You, and Yu]{zhang2006automatic}
Chi Zhang, Zheng You, and Shijie Yu.
\newblock An automatic exposure algorithm based on information entropy.
\newblock In \emph{Sixth International Symposium on Instrumentation and Control
  Technology: Signal Analysis, Measurement Theory, Photo-Electronic Technology,
  and Artificial Intelligence}, pages 152--156. SPIE, 2006.

\bibitem[Zhang et~al.(2020)Zhang, Tian, Kong, Zhong, and Fu]{zhang2020residual}
Yulun Zhang, Yapeng Tian, Yu Kong, Bineng Zhong, and Yun Fu.
\newblock Residual dense network for image restoration.
\newblock \emph{IEEE Transactions on Pattern Analysis and Machine
  Intelligence}, 43\penalty0 (7):\penalty0 2480--2495, 2020.

\bibitem[Zhang et~al.()Zhang, Xu, Liu, Yan, and Zuo]{zhang2022self}
Zhilu Zhang, RongJian Xu, Ming Liu, Zifei Yan, and Wangmeng Zuo.
\newblock Self-supervised image restoration with blurry and noisy pairs.
\newblock In \emph{Advances in Neural Information Processing Systems}.

\bibitem[Zhong et~al.(2020)Zhong, Gao, Zheng, and Zheng]{zhong2020efficient}
Zhihang Zhong, Ye Gao, Yinqiang Zheng, and Bo Zheng.
\newblock Efficient spatio-temporal recurrent neural network for video
  deblurring.
\newblock In \emph{Computer Vision--ECCV 2020: 16th European Conference,
  Glasgow, UK, August 23--28, 2020, Proceedings, Part VI 16}, pages 191--207.
  Springer, 2020.

\bibitem[Zhong et~al.(2022)Zhong, Gao, Zheng, Zheng, and Sato]{zhong2022real}
Zhihang Zhong, Ye Gao, Yinqiang Zheng, Bo Zheng, and Imari Sato.
\newblock Real-world video deblurring: A benchmark dataset and an efficient
  recurrent neural network.
\newblock 2022.

\end{thebibliography}
}


\end{document}